\documentclass[sigconf,authordraft,submit,review=false]{acmart}
\usepackage{multirow}

\preto{\abstractkeywords}{\nolinenumbers}

\usepackage{booktabs}
\usepackage{graphicx}
\usepackage{makecell}
\usepackage{array}
\usepackage{adjustbox}

\usepackage{pifont}
\newcommand{\cmark}{\ding{51}} 
\newcommand{\xmark}{\ding{55}} 
\AtBeginDocument{%
  }

\setcopyright{acmlicensed}
\copyrightyear{2026}
\acmYear{2026}
\acmDOI{XXXXXXX.XXXXXXX}
\acmConference[ACM MM 2026]{ACM International Conference on Multimedia}{November 10--14, 2026}{Brazil}
\acmISBN{978-1-4503-XXXX-X/2018/06}

\begin{document}

\title[MARINER]{MARINER: A 3E-Driven Benchmark for Fine-Grained Perception and Complex Reasoning in Open-Water Environments}


\author{Xingming Liao}
\email{liaoxingming@mails.gdut.edu.cn}
\orcid{0009-0000-9118-5968}
\affiliation{%
  \institution{Guangdong University of Technology}
  \city{Guangzhou}
  \country{China}
}

\author{Ning Chen}
\email{18162195164@163.com}
\affiliation{%
  \institution{Guangdong University of Technology}
  \city{Guangzhou}
  \country{China}
}

\author{Muying Shu}
\email{shumuying8@gmail.com}
\affiliation{%
  \institution{Guangdong University of Technology}
  \city{Guangzhou}
  \country{China}
}

\author{YunPeng Yin}
\email{yypeng@mail2.gdut.edu.cn}
\affiliation{%
  \institution{Guangdong University of Technology}
  \city{Guangzhou}
  \country{China}
}

\author{Peijian Zeng}
\email{lil_ken@163.com}
\affiliation{%
  \institution{Guangdong University of Technology}
  \city{Guangzhou}
  \country{China}
}

\author{Zhuowei Wang}
\authornote{Zhuowei Wang and Nankai Lin are co-corresponding authors.}
\email{zwwang@gdut.edu.cn}
\affiliation{%
  \institution{Guangdong University of Technology}
  \city{Guangzhou}
  \country{China}
}

\author{Nankai Lin}
\authornotemark[1]
\email{neakail@outlook.com}
\affiliation{%
  \institution{Guangdong University of Technology}
  \city{Guangzhou}
  \country{China}
}
\author{Lianglun Cheng}
\email{llcheng@gdut.edu.cn}
\affiliation{%
  \institution{Guangdong University of Technology}
  \city{Guangzhou}
  \country{China}
}

\renewcommand{\shortauthors}{Xingming Liao et al.}

\begin{abstract}
Fine-grained visual understanding and high-level reasoning in real-world open-water environments remain under-explored due to the lack of dedicated benchmarks. We introduce MARINER, a comprehensive benchmark built under the novel \textbf{E}ntity-\textbf{E}nvironment-\textbf{E}vent \textbf{(3E)} paradigm. MARINER contains 16,629 multi-source maritime images with 63 fine-grained vessel categories, diverse adverse environments, and 5 typical dynamic maritime incidents, covering fine-grained classification, object detection, and visual question answering tasks. We conduct extensive evaluations on mainstream Multimodal Large language models (MLLMs) and establish baselines, revealing that even advanced models struggle with fine-grained discrimination and causal reasoning in complex marine scenes. As a dedicated maritime benchmark, MARINER fills the gap of realistic and cognitive-level evaluation for maritime multimodal understanding, and promotes future research on robust vision-language models for open-water applications. Appendix and supplementary materials are available at https://lxixim.github.io/MARINER.
\end{abstract}

\ccsdesc[500]{Computing methodologies~Computer vision}

\keywords{Fine-grained Visual Understanding, Maritime Perception and Reasoning, Benchmark, Multimodal Large language models}



\maketitle

\begin{table*}[t]
\centering
\caption{Comparison of ship-related datasets in terms of source diversity, category scale, environmental coverage, event representation, task coverage, and dataset scale. Cls., Det., and VQA denote classification, detection, and visual question answering, respectively. Multi-source denotes that the dataset is collected from open-source imagery, self-developed electro-optical pod platform and unmanned aerial vehicle.}
\label{tab:datasets}
\setlength{\tabcolsep}{4pt}
\begin{tabular}{l p{2.2cm} c c c c c c c c}
\toprule
\textbf{Name} & \textbf{Source} & \textbf{Categories} & \textbf{Environment} & \textbf{Event} & \textbf{Cls.} & \textbf{Det.} & \textbf{VQA} & \textbf{Images} & \textbf{Instances} \\
\midrule
MS COCO \cite{lin2014microsoft}   & Open-source   & 1  & Limited  & Static  & \xmark & \cmark & \xmark & 3,164  & 11,189  \\
SeaShip \cite{shao2018seaships}   & Shore-based   & 6  & Limited  & Static  & \cmark & \cmark & \xmark & 31,455 & 40,077  \\
Boat Re-ID \cite{spagnolo2019new} & Shore-based   & - & Limited  & Static  & \cmark & \xmark & \xmark & 5,523  & 5,523   \\
McShips \cite{zheng2020mcships}   & Open-source   & 13 & Diverse  & Static  & \cmark & \cmark & \xmark & 14,709 & 26,529  \\
ABOships \cite{iancu2021aboships} & Shipboard     & 9  & Limited  & Limited & \cmark & \cmark & \xmark & 9,880  & 41,967  \\
SeaSAw \cite{kaur2022sea}         & Shipboard     & 12 & Diverse  & Dynamic & \cmark & \cmark & \xmark & 1.9M   & 14.6M   \\
IFShip \cite{guo2025ifship}       & Open-source   & 17 & Moderate & Static  & \cmark & \xmark & \cmark & 18,929 & -  \\
\midrule
\textbf{MARINER (Ours)} & \textbf{Multi-source} & \textbf{63} & \textbf{Diverse} & \textbf{Dynamic} & \textbf{\cmark} & \textbf{\cmark} & \textbf{\cmark} & \textbf{16,629} & \textbf{17,895} \\
\bottomrule
\end{tabular}
\end{table*}

\section{Introduction}

\begin{figure}[ht]
  \centering
   \includegraphics[width=0.75\linewidth]{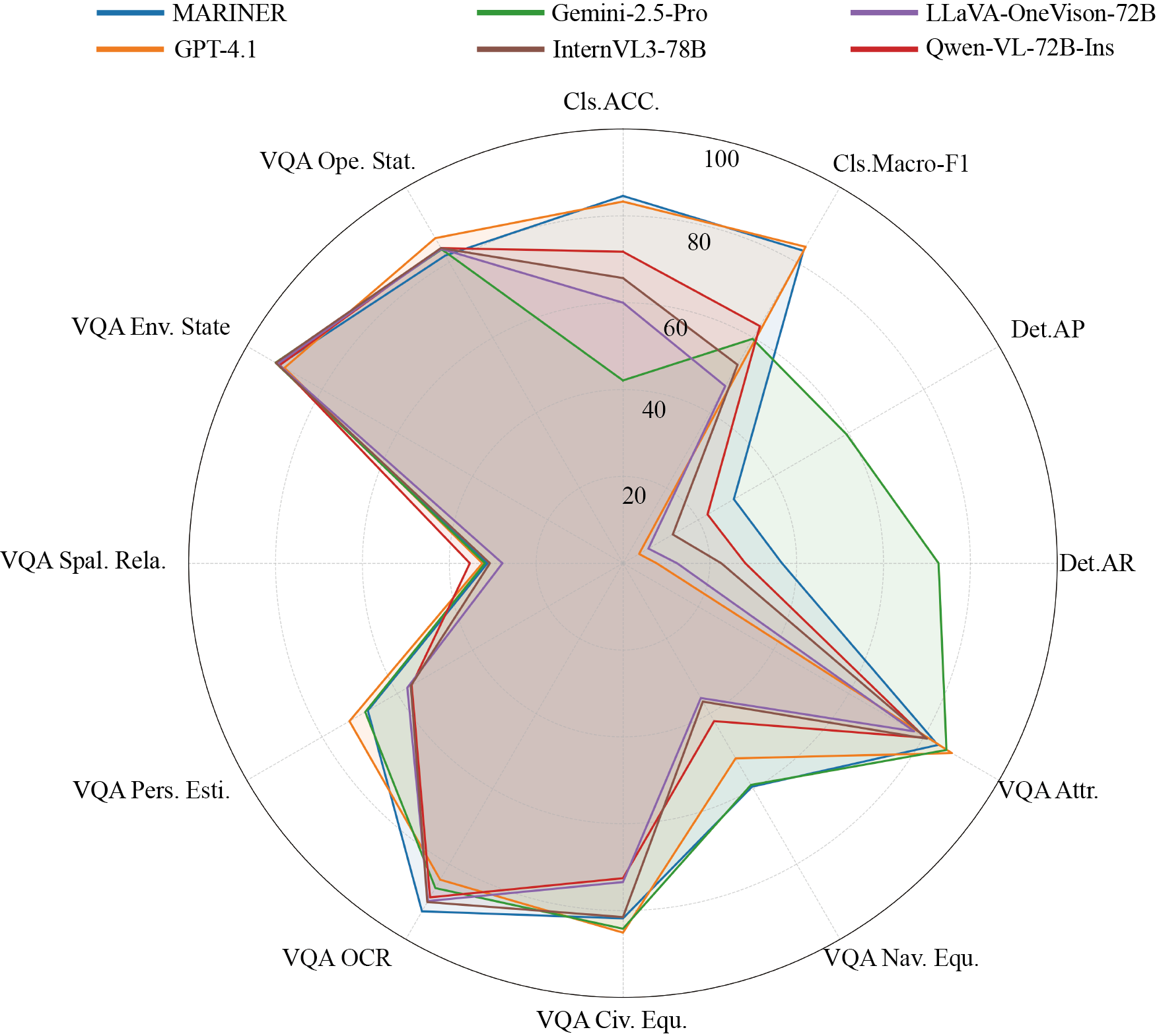}
   \caption{Normalized performance comparison of MARINER and other models \cite{hurst2024gpt,qwen2025qwen25technicalreport,li2024llava,zhu2025internvl3,comanici2025gemini} across the key metrics of the three benchmark tasks.}
   \label{fig:all_radar}
\end{figure}

The maritime domain is of growing importance for numerous civil and strategic applications. With the rapid advancement of autonomous shipping, maritime intelligent systems require fine-grained visual understanding and reasoning over complex open-water scenes. Unlike controlled laboratory environments, real-world maritime scenarios suffer from extreme weather, complex vessel interactions, and emergent incidents, posing unprecedented challenges for robust visual understanding and decision-making.

Recent advances in Large Language Models (LLMs) \cite{brown2020language,touvron2023llama} and Multimodal Large Language Models (MLLMs) \cite{bai2025qwen3,team2024gemini} have opened new possibilities for intelligent maritime perception and reasoning. These models have demonstrated strong capabilities in structured visual understanding, generation \cite{chen2021evaluating,clouatre2019figr}, and visual question answering (VQA) \cite{li2022blip}, and are increasingly applied to domain-specific complex visual tasks \cite{LIAO2025103134,wang2025embodied,chen2025mimo}. However, a key challenge remains: \textbf{how to systematically assess model capabilities required for autonomous navigation and intelligent maritime management in realistic open-water scenarios?} Existing benchmark tests have three key limitations: (1) \textbf{Limited granularity:} ship taxonomies are insufficient to capture the subtle semantic distinctions required for fine-grained vessel understanding; (2) \textbf{Limited realism:} overly simplified maritime scenes cannot capture the environmental and event complexity of real open-water settings; (3) \textbf{Evaluation fragmentation:} it evaluates individual tasks in isolation, rather than the integrated perception, spatial understanding, and reasoning capabilities required for practical maritime applications. Existing maritime benchmarks can be broadly divided into three streams: remote-sensing-based perception \cite{chen2022degraded,chen2020fgsd,gallego2018automatic,li2020object}, camera-based perception \cite{su2023survey,zhang2023unsupervised,kaur2022sea,spagnolo2019new,lin2014microsoft}, and multimodal maritime understanding \cite{guo2025ifship}. The first two predominantly address perception-level tasks such as classification and detection, whereas the third moves toward cognition-oriented maritime understanding through VQA. However, current benchmarks lack the granularity and realism required for realistic open-water understanding, particularly in fine-grained vessel recognition, diverse environmental conditions, and maritime event understanding.

To address these limitations, we introduce MARINER, a comprehensive benchmark for fine-grained maritime understanding in real-world open-water environments. Based on an \textbf{E}ntity-\textbf{E}nvironment-\textbf{E}vent \textbf{(3E)} paradigm, MARINER covers fine-grained vessel categories, diverse maritime environments, and dynamic incident scenarios, offering a more realistic and structured representation of real-world maritime complexity. The benchmark supports three downstream tasks, namely fine-grained classification, fine-grained object detection, and VQA, enabling systematic evaluation from low-level perception to high-level reasoning, as further summarized in the radar-chart comparison shown in Figure~\ref{fig:all_radar}. In particular, the VQA component is organized along multiple dimensions, including perception, spatial understanding, and reasoning, allowing a more holistic assessment of the capabilities required for autonomous navigation and intelligent maritime management.
Our contributions can be summarized as follows: 1) We fill a critical gap in existing maritime benchmarks for maritime intelligent systems by capturing realistic open-water complexity from 3E paradigm. 2) We introduce MARINER, a large-scale maritime benchmark that supports the progression from detection to perception and further to understanding through unified classification, detection, and VQA tasks. 3) We provide extensive evaluation of MLLMs on MARINER, establishing performance baselines and offering insights into their capabilities and limitations in realistic maritime scenarios.

\section{Related Work}

\subsection{Multimodal Large Language Models}
Recent years have witnessed rapid progress in MLLMs \cite{qwen2025qwen25technicalreport,NEURIPS2023_6dcf277e,chen2021evaluating,li2024survey}, which have shown strong performance on a broad range of general-domain tasks, such as visual understanding \cite{li2022blip}, object grounding \cite{huang2025high,peng2023kosmos}, and VQA \cite{cheng2025simplevqa,wang2025traceable}. Encouraged by these successes, many studies have explored the adaptation of MLLMs to domain-specific scenarios \cite{jiang2024mmad,huang2025towards}, highlighting their potential for complex visual analysis and knowledge-intensive reasoning. The maritime domain is a representative scenario for such capabilities, as autonomous navigation \cite{cui2024survey,salgado2026usv} and intelligent maritime management \cite{zhang2025multilingual} require models to jointly support fine-grained vessel recognition, target localization, environmental perception, and reasoning about operational states and maritime incidents in complex open-water scenes. As a result, maritime scenarios impose higher requirements on the integrated capabilities of MLLMs.

\begin{figure*}[t]
  \includegraphics[width=0.85\textwidth]{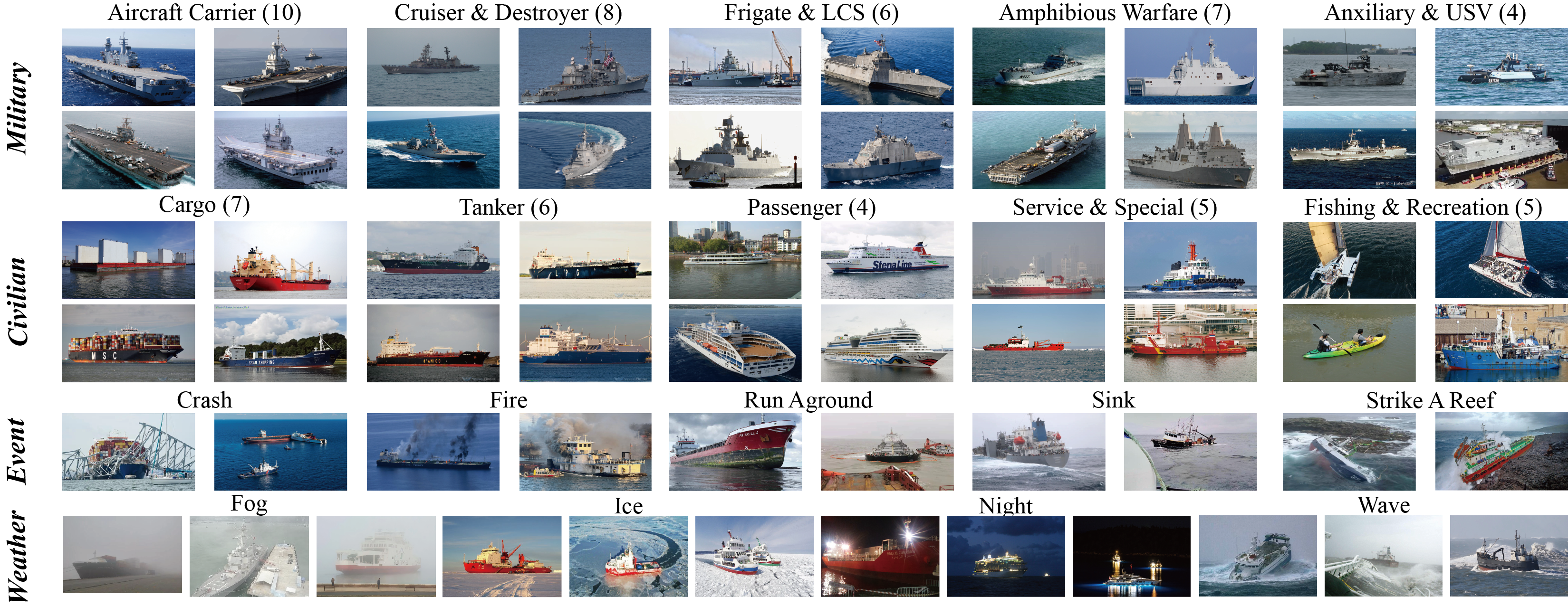}
  \caption{Representative examples from the MARINER dataset, covering fine-grained military and civilian ship categories, maritime incident scenarios, and adverse environmental conditions. The number in parentheses after each group name denotes the number of corresponding categories included in that group.}
  \label{fig:case_ship}
\end{figure*}

\subsection{Maritime Benchmarks}
An increasing number of maritime benchmarks have been developed for ship perception and evaluation. Remote-sensing benchmarks \cite{chen2022degraded,zhang2021shiprsimagenet,chen2020fgsd,gallego2018automatic,li2020object} mainly focus on ship detection and classification in aerial or satellite imagery, whereas camera-based benchmarks \cite{zheng2020mcships,spagnolo2019new,shao2018seaships,lin2014microsoft} emphasize near-field perception for applications such as port monitoring and coastal surveillance. These benchmarks have enriched maritime evaluation from diverse imaging perspectives and practical application settings. However, they are still predominantly oriented toward perception-level tasks, especially detection and classification. Recent maritime benchmarks \cite{guo2025ifship,wang2025cot4ad} have expanded evaluation toward higher-level understanding. Nevertheless, current researches still lack a benchmark for assessing whether models can meet the demands of autonomous navigation and intelligent maritime management in realistic open-water scenarios. This gap is reflected in three key aspects: limited granularity, limited realism, and fragmented evaluation. To bridge this gap, we propose MARINER, a unified benchmark for fine-grained maritime understanding that integrates vessel categories, maritime environments, and incident scenarios.

\section{MARINER}
\subsection{Data Collection}
We construct a large-scale multi-task MARINER benchmark, targeting fine-grained ship understanding in realistic maritime settings. To model open-water visual complexity, we drive our collection process using 3E paradigm as shown in Figure~\ref{fig:case_ship}:

\textbf{1) Entity.} We establish a rigorous hierarchical taxonomy of 63 categories, encompassing 35 military classes, 27 civilian classes, and an ``other'' category for ambiguous or unidentifiable instances, as illustrated in Figure~\ref{fig:category}. More detailed category definitions and statistical analyses are provided in the Appendix B.

\textbf{2) Environment.} To capture the severe visual degradation typical of real-world deployments, we sample images across four adverse conditions: fog, ice, night, and waves. Different weather conditions are introduced to test the model robustness in real scenarios.

\begin{figure}[htbp]
  \centering
   \includegraphics[width=0.8\linewidth]{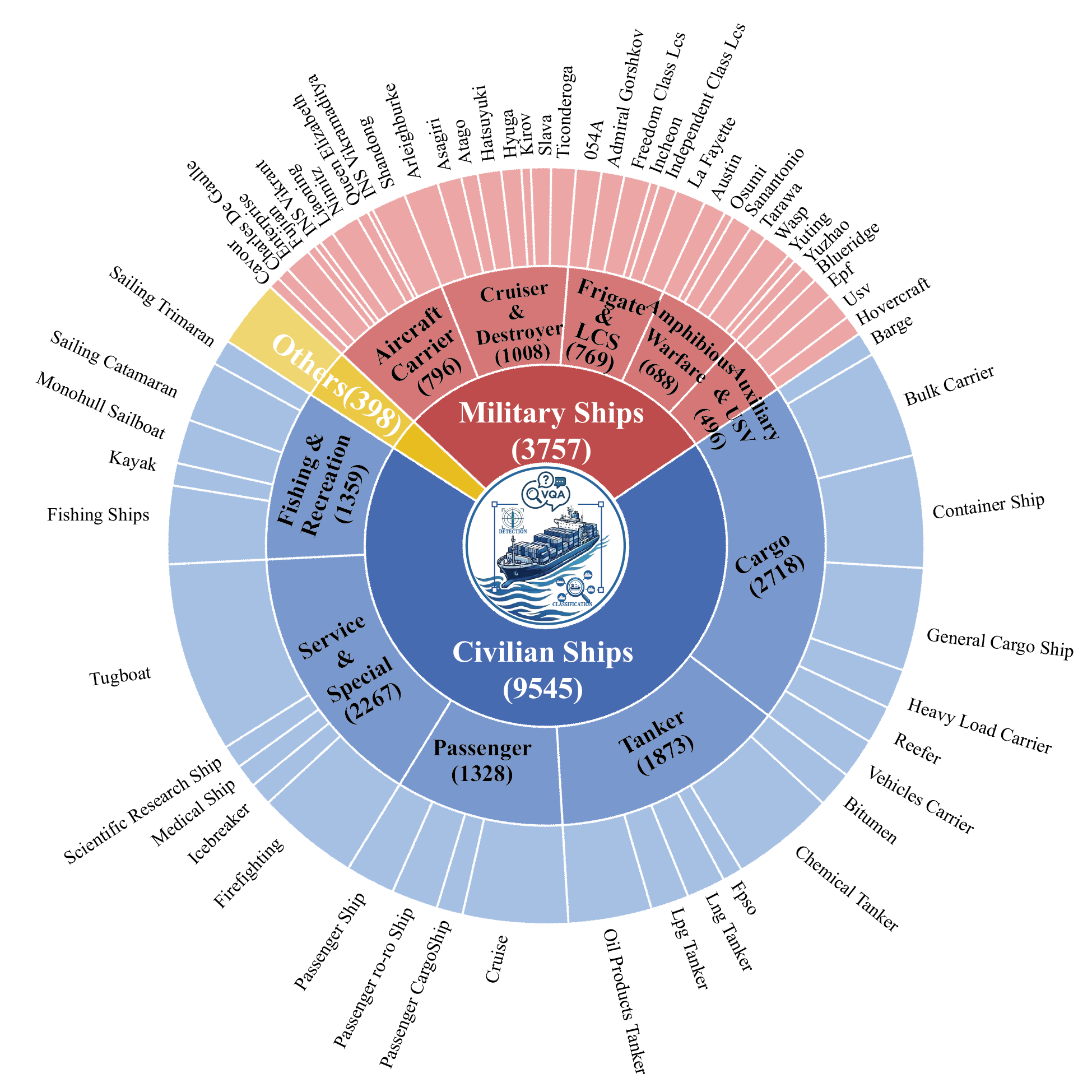}
   \caption{Distribution of the annotated instances within the MARINER,
formulated specifically for the fine-grained classification and fine-grained detection tasks.}
   \label{fig:category}
\end{figure}

\textbf{3) Event.} We incorporate five critical maritime incidents: crash, fire, run aground, sink, and strike a reef. This dimension broadens the benchmark toward event-centric and state-aware understanding, supporting incident interpretation and event-level reasoning.

We initialize the foundational corpus by collecting extensive open-source maritime video and image archives. To incorporate operational realism and capture multi-view dynamics, we augment this baseline with raw sensor streams extracted directly from spherical electro-optical payloads and UAV. Furthermore, to capture rare dynamic maritime events, we explicitly extract hard examples from official registries, multinational accident databases, and global news broadcasts. Following rigorous data collection, filtering, and quality control, we obtain 16,629 high-quality maritime images. Both classification/detection benchmark and the VQA benchmark are derived from this shared corpus, but they are built from overlapping rather than identical image subsets.

\subsection{Data Annotation and VQA Pairs Generation}
\noindent \textbf{Fine-Grained Ship Annotation.}
We implement a human-in-the-loop annotation pipeline that couples foundation-model localization with prior-driven label propagation and expert refinement. 

\begin{figure*}[ht]
  \centering
   \includegraphics[width=0.8\linewidth]{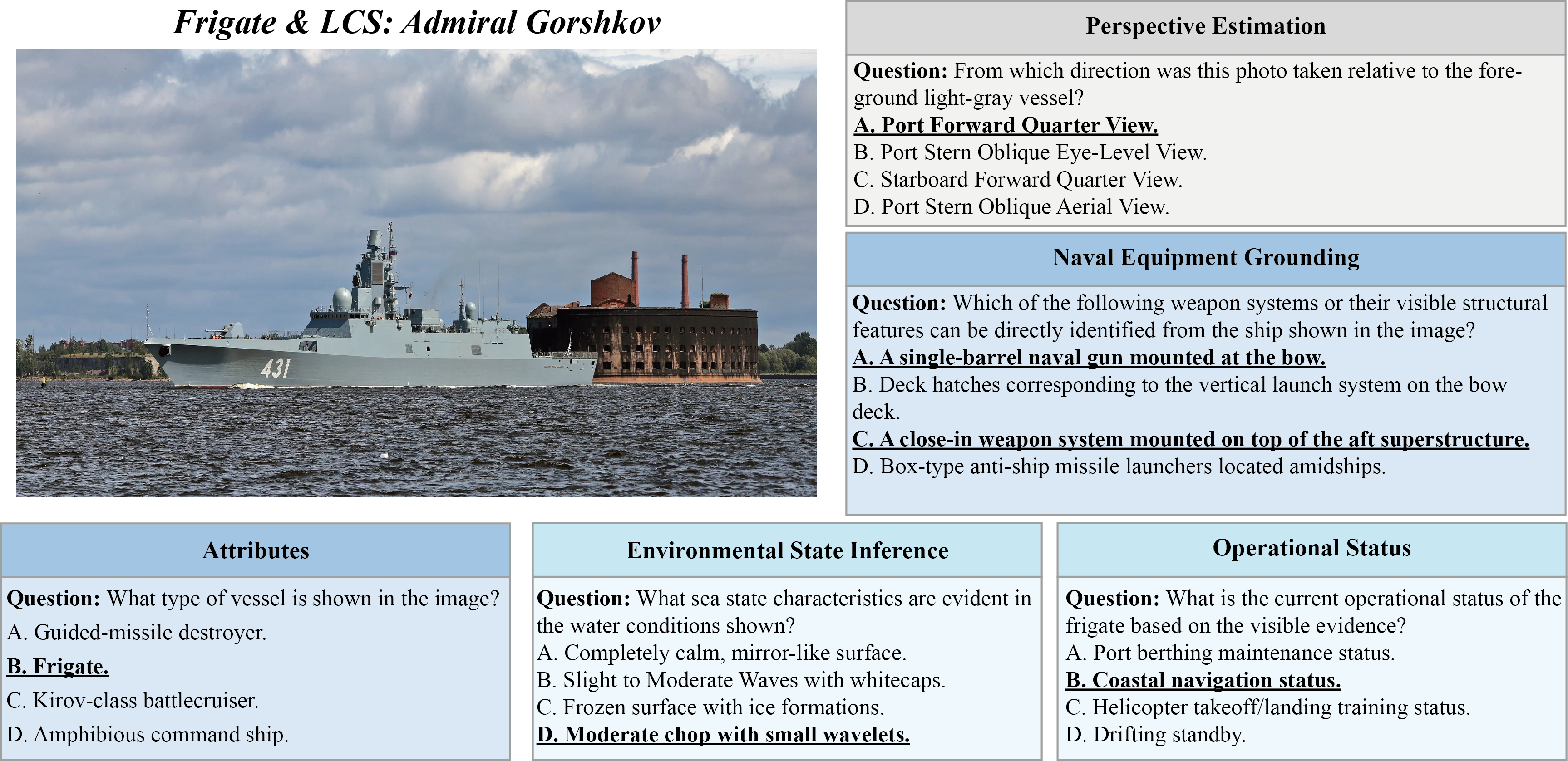}
   \caption{Example of a partially instantiated VQA sample from MARINER. Although the benchmark is designed with eight question categories, not every image can support all of them. The applicable question types are determined by the visible objects, scene composition, and reasoning cues available in each image. The figure shows several representative question-answer pairs for this image, where the bold underlined option denotes the ground-truth answer.}
   \label{fig:vqa_case}
\end{figure*}

\noindent \textit{Stage 1: Class-Agnostic Localization.} Since vision foundation models (e.g., YOLO \cite{varghese2024yolov8} and Grounding DINO \cite{liu2024grounding}) are not reliable for fine-grained maritime categorization, we use them to detect ship instances in a class-agnostic manner, labeling all vessels as \textit{ship}. To maximize recall, we merge predictions from multiple models using Non-Maximum Suppression with an IoU threshold of 0.5.

\noindent \textit{Stage 2: Prior-Driven Label Propagation.} Exploiting the strict taxonomic organization of our image corpus, we leverage folder-level semantics as priors. Class-agnostic boxes are automatically assigned the label of their parent directory. This top-down propagation circumvents the bottleneck of instance-level classification, yielding efficient and structurally consistent preliminary annotations.

\noindent \textit{Stage 3: Quality Control.} To enforce benchmark rigor, domain experts execute an exhaustive review of the initial annotations, targeting four critical dimensions: (1) tightening spatial boundaries around the hull; (2) recovering false negatives (e.g., heavily obscured or tiny vessels) missed by foundation models; (3) pruning false positives triggered by background distractors (e.g., breaking waves, port infrastructure); and (4) rectifying semantic misassignments in multi-class scenes. Unidentifiable instances are forcefully reassigned to the ``Other'' class. Following an independent cross-validation phase, the final benchmark generates a corpus of 11,790 images encompassing 13,700 precisely annotated ship instances.


\noindent \textbf{VQA Task Definition.}
To systematically benchmark MLLMs, we structure the eight VQA dimensions across three cognitive processing levels: Perception, Space, and Reasoning:

\textbf{1) Perception} requires models to accurately extract and ground fine-grained structural semantics from raw visual input. Confronting open-water visual complexity, this level rigorously tests cross-modal alignment and the localized recognition of explicit physical attributes, establishing the baseline for visual comprehension.

\begin{itemize}
    \item \textbf{Attributes:} Probes the extraction of explicit physical characteristics, including precise vessel classification, hull color, superstructure layout, and functional mission attributes from local visual cues.

    \item \textbf{Naval Equipment Grounding:} Targets the localization of military payloads, such as close-in weapon systems, missile launchers, and carrier-based aircraft, demanding expert-level semantic matching within complex tactical layouts.

    \item \textbf{Civilian Equipment Grounding:} Challenges the model to detect diverse commercial components (e.g., high-pedestal cranes, lifeboats) amidst highly homogeneous civilian hull designs, resolving severe inter-class visual ambiguities.

    \item \textbf{OCR-Integrated QA:} Evaluates scene text recognition by extracting critical operational identifiers (e.g., painted ship names, IMO numbers) subjected to perspective distortion or physical degradation.
\end{itemize}

\textbf{2) Space} challenges the model to reconstruct 3D configurations and topological layouts from 2D projections. In unpredictable maritime environments, this dimension tests spatial awareness, demanding accurate interpretation of camera perspectives, relative navigational positioning, and multi-object geometric orientations.

\begin{itemize}
    \item \textbf{Perspective Estimation:} Requires reconstructing 3D geometric orientations (e.g., frontal bow, oblique quarter) by interpreting structural cues like vanishing points, deck visibility, and hull foreshortening to answer questions dependent on the vessel's global spatial posture.

    \item \textbf{Spatial Relationship:} Tests the resolution of inter-object direction, viewer-centric orientations, and multi-vessel heading alignments using dual spatial reference frames to answer questions requiring complex navigational alignment.
\end{itemize}

\textbf{3) Reasoning} forces models to transition from isolated object recognition to complex event inference by synthesizing global multi-source cues. Confronting unpredictable maritime incidents, this highest cognitive level requires models to interpret physical logic and causal interactions (e.g., linking water wakes to vessel speed) to deduce ongoing operational states and environmental hazards.

\begin{itemize}
    \item \textbf{Environmental State Inference:} Demands the deduction of meteorological and geographical conditions (e.g., hazardous sea states, visibility degradation, and emergency operations) from global visual textures and lighting variations.

    \item \textbf{Operational Status:} Focuses on disambiguating mutually exclusive scenarios (e.g., normal navigation vs. critical grounding/fire) by analyzing implicit causal markers like water wakes, smoke trajectories, and hull listing angles to answer questions dependent on precise dynamic state identification.
\end{itemize}

\noindent \textbf{VQA Pairs Generation.} 
We introduce a constraint-driven generation pipeline that bridges deterministic visual metadata with the generative reasoning capabilities of MLLMs.

\noindent \textit{Stage 1: Domain Knowledge and Deterministic Constraint Formulation.} To construct a factual baseline, we aggregate unstructured data from Wikipedia and specialized maritime registries. We prompt MLLMs to distill these texts into strictly visual schemas, stripping unobservable trivia (e.g., historical timelines). To prevent MLLMs' spatial hallucinations, we establish deterministic metadata anchors: extracting explicit bounding boxes, calculating coordinate offsets, and mapping text strings to exact hull locations. For reasoning tasks, we introduce a \textit{Dual-Fact Pool} mechanism. The actual visual state defines the Ground Truth Pool, while mutually exclusive states populate the Distractor Pool, establishing rigid boundaries for downstream generation.

\noindent \textit{Stage 2: Iterative Constraint-Driven Generation.} Using these deterministic information, we execute an adversarial Generate-Check-Regenerate pipeline. During \textit{Generation}, Question-Option-Answer triplets grounded in the Dual-Fact Pools are formulated by an MLLM. During \textit{Validation}, an independent MLLMs instance acts as an automated evaluator, screening for ungrounded visual hallucinations, linguistic leakage, and logical overlaps. Flawed triplets are pushed to a \textit{Refinement} phase for up to three self-correction loops. Unresolvable triplets are permanently discarded.

\noindent \textit{Stage 3: Quality Control.} In the final phase, independent maritime annotators manually audit the surviving triplets derived from 13,397 images, eliminating residual semantic ambiguities, verifying alignment between visual evidence and answers, and ensuring that all distractors remain plausible under maritime constraints. This process yields a final dataset comprising 33,125 validated QA pairs.  Representative examples are shown in Figure~\ref{fig:vqa_case}.

\section{Experiments}

\begin{table}[t]
\centering
\caption{Classification performance comparison of different models. Best performances for open-source models are highlighted in bold.}
\label{tab:classification_results}
\resizebox{\columnwidth}{!}{%
\begin{tabular}{lccccc}
\toprule
\multirow{2}{*}{Model} & \multirow{2}{*}{Scale} & \multicolumn{4}{c}{Classification} \\ 
\cmidrule(lr){3-6}
& & Accuracy & Macro F1 & mAcc & Weighted F1 \\
\midrule
Human (expert)         & -     & 82.52      & -      & -      & -      \\
\midrule
GPT-4o                 & -     & 80.50  & 80.89  & 81.01  & 80.90  \\
GPT-4.1                & - & 83.30  & 84.14 & 84.50 & 82.77    \\
Gemini-2.5-Flash       & -     & 82.60  & 80.36  & 81.73  & 81.79  \\
Gemini-2.5-Pro         & -     & 42.10  & 59.68  & 59.68  & 42.10  \\
\midrule
Qwen2.5-VL-Ins      & 3B    & 52.60  & 37.16  & 42.53  & 47.74  \\
LLaVA-1.5-hf           & 7B    & 14.50  & 7.52   & 6.53   & 16.65  \\
Qwen2.5-VL-Ins         & 7B    & 64.40  & 51.18  & 57.45  & 60.40  \\
LLaVA-OneVision        & 7B    & 41.20  & 28.14  & 33.56  & 34.43  \\
InternVL2           & 8B    & 26.30  & 13.67  & 15.86  & 24.75  \\
InternVL3           & 8B    & 41.00  & 22.03  & 26.40  & 34.99  \\
MiniCPM-V-2.6          & 8B    & 49.00  & 35.92  & 39.86  & 44.77  \\
Qwen2.5-VL-Ins         & 32B   & 65.80  & 59.22  & 62.01  & 62.28  \\
InternVL3          & 38B   & 60.10  & 49.37  & 50.65  & 56.66  \\
Qwen2.5-VL-Ins         & 72B   & 71.75  & 63.06  & 68.95  & 68.71  \\
LLaVA-OneVision       & 72B     & 60.00      & 47.12      & 50.97      & 56.75      \\
InternVL3          & 78B   & 65.67  & 52.78  & 56.96  & 62.84  \\
\midrule
MARINER                & 7B    & \textbf{84.60} & \textbf{83.10} & \textbf{82.62} & \textbf{84.67} \\
$\Delta$ \textit{v.s.} Qwen2.5-VL-Ins
                       & 7B     & $\uparrow$ 20.20 & $\uparrow$ 31.92 & $\uparrow$ 25.17 & $\uparrow$ 24.27 \\
\bottomrule
\end{tabular}%
}
\end{table}

\begin{table}[t]
\centering
\caption{Detection performance comparison of different models. Best performances for open-source models are highlighted in bold. GPT-series models do not report results on military vessel categories, due to safety constraints related to military-target recognition.}
\label{tab:detection_results}
\resizebox{\columnwidth}{!}{%
\begin{tabular}{lc cccc|cc}
\toprule
\multirow{2}{*}{Model} & \multirow{2}{*}{Scale} & \multicolumn{6}{c}{Detection} \\ 
\cmidrule(lr){3-8}
& & AP & AP$_{50}$ & AP$_{75}$ & AR & AP$_{\text{Mil}}$ & AP$_{\text{Civ}}$ \\
\midrule
Human (expert)    & -  & 68.30 & - & - & - & 49.12 & 87.48 \\
\midrule
GPT-4o (Civ)            & -  & 1.46 & 6.06 & 0.34 & 3.12 & - & 3.12 \\
GPT-4.1 (Civ)        & -  & 4.28 & 12.98 & 1.75 & 7.75 & - & 9.14 \\
Gemini-2.5-Flash  & -  & 16.94 & 35.67 & 13.72 & 24.34 & 22.91 & 10.14 \\
Gemini-2.5-Pro    & -  & 59.42 & 68.46 & 60.62 & 72.68 & 69.04 & 48.47 \\
\midrule
Qwen2.5-VL-Ins    & 3B  & 4.88 & 8.37 & 4.93 & 7.74 & 8.35 & 0.94 \\
Qwen2.5-VL-Ins    & 7B  & 16.42 & 19.65 & 16.40 & 21.90 & 26.10 & 5.40 \\
LLaVA-OneVision   & 7B  & 6.11 & 13.11 & 5.00 & 9.41 & 7.59 & 4.43 \\
InternVL2         & 8B  & 1.07 & 3.27 & 0.46 & 3.16 & 0.43 & 1.81 \\
InternVL3         & 8B  & 6.82 & 11.83& 6.58 & 11.16 & 2.37 & 11.87 \\
MiniCPM-V-2.6     & 8B  & 9.18 & 20.79 & 5.96 & 13.38 & 5.93 & 12.87 \\
Qwen2.5-VL-Ins    & 32B & 17.13 & 20.60 & 17.05 & 22.70 & 26.91 & 6.00 \\
InternVL3         & 38B & 27.06 & 33.70 & 28.92 & \textbf{37.05} & 21.77 & 33.07 \\
Qwen2.5-VL-Ins    & 72B & 22.49 & 26.24 & 22.33 & 28.17 & \textbf{37.29} & 5.64 \\
LLaVA-OneVision   & 72B & 6.76 & 18.47 & 4.41 & 12.46 & 7.35 & 6.08 \\
InternVL3         & 78B & 13.25 & 27.91 & 9.85 & 22.64 & 7.91 & 19.32 \\
\midrule
MARINER           & 7B  & \textbf{29.49} & \textbf{44.32} & \textbf{31.80} & 36.62 & 23.83 & \textbf{35.93} \\
$\Delta$ \textit{v.s.} Qwen2.5-VL-Ins 
                 & 7B & $\uparrow$ 13.07  & $\uparrow$ 24.67 & $\uparrow$ 15.40 & $\uparrow$ 14.72 & $\downarrow$ 2.27 & $\uparrow$ 30.53  \\
\bottomrule
\end{tabular}
}
\end{table}

\begin{table*}[t]
\centering
\small
\caption{Results of different models on various task types in the VQA benchmark. The evaluation is reported across three major dimensions, including Perception, Space, and Reasoning. Best performances for open-source models are highlighted in bold.}
\label{tab:task_group_results}
\begin{tabular}{l c cccccccc}
\toprule
& & \rotatebox{60}{Attributes} 
& \rotatebox{60}{Nav. Equ.} 
& \rotatebox{60}{Civ. Equ.} 
& \rotatebox{60}{OCR} 
& \rotatebox{60}{Pers. Esti.} 
& \rotatebox{60}{Spal. Rela.} 
& \rotatebox{60}{Env. State} 
& \rotatebox{60}{Ope. Stat.} \\
\cmidrule(lr){3-6} \cmidrule(lr){7-8} \cmidrule(lr){9-10}
\multirow{-2}{*}{Model} & \multirow{-2}{*}{Overall} 
& \multicolumn{4}{c}{Perception} 
& \multicolumn{2}{c}{Space} 
& \multicolumn{2}{c}{Reasoning} \\
\midrule
Human (expert) & 82.63 & 55.30 & 44.75 & 80.45 & 99.85 & 99.80 & 99.72 & 89.10 & 92.03 \\
\midrule
GPT-4o & 68.91 & 82.97 & 47.64 & 81.19 & 78.22 & 62.70 & 35.26 & 85.32 & 77.97 \\
GPT-4.1 & 73.77 & 87.46 & 51.89 & 85.07 & 84.16 & 72.73 & 32.37 & 90.01 & 86.45 \\
Gemini-2.5-Flash & 64.52 & 75.09 & 45.28 & 68.96 & 58.42 & 59.67 & 33.53 & 90.08 & 85.13 \\
Gemini-2.5-Pro & 73.97 & 86.02 & 58.96 & 84.18 & 86.39 & 68.53 & 31.79 & 92.38 & 83.48 \\
\midrule
Qwen2.5-VL-3B-Ins & 60.67 & 69.21 & 33.02 & 60.00 & 85.89 & 39.16 & 30.64 & 89.59 & 77.86 \\
LLaVA-1.5-7B-hf & 49.86 & 53.88 & 25.94 & 49.85 & 66.58 & 31.93 & 22.54 & 79.25 & 68.94 \\
Qwen2.5-VL-7B-Ins & 65.63 & 74.26 & 35.38 & 73.73 & 88.61 & 51.52 & 27.75 & 91.33 & 82.49 \\
LLaVA-OneVision-7B & 60.47 & 72.52 & 23.11 & 67.16 & 83.17 & 41.96 & 24.86 & 89.03 & 81.94 \\
InternVL2-8B & 60.08 & 63.81 & 21.23 & 64.48 & 88.61 & 41.72 & 31.79 & 88.61 & 80.40 \\
InternVL3-8B & 64.74 & 71.56 & 27.83 & 77.01 & 87.38 & 50.58 & 29.48 & 90.85 & 83.26 \\
MiniCPM-V-2.6-8B & 61.57 & 68.51 & 23.11 & 68.06 & 91.58 & 44.29 & 26.01 & 90.78 & 80.18 \\
Qwen2.5-VL-32B-Ins & 63.85 & 75.96 & 28.77 & 74.63 & 89.11 & 49.65 & \textbf{35.26} & 74.63 & 82.82 \\
InternVL3-38B & 69.29 & 80.36 & 35.38 & \textbf{82.09} & 89.85 & 55.94 & 33.53 & 91.82 & \textbf{85.35} \\
Qwen2.5-VL-72B-Ins & 68.79 & 80.40 & 41.98 & 72.54 & 88.86 & 56.18 & \textbf{35.26} & 91.26 & 83.81 \\
LLaVA-OneVision-72B & 67.13 & 77.44 & 35.85 & 73.43 & 89.85 & 57.34 & 27.75 & 91.82 & 83.59 \\
InternVL3-78B & 69.05 & 80.84 & 36.79 & 81.49 & 90.10 & 56.41 & 30.64 & \textbf{92.31} & 83.81 \\
\midrule
MARINER & \textbf{73.72} & \textbf{83.71} & \textbf{59.43} & 81.79 & \textbf{92.57} & \textbf{67.83} & 31.21 & 91.40 & 81.83 \\
$\Delta$ \textit{v.s.} Qwen2.5-VL-7B & $\uparrow$ 8.09 & $\uparrow$ 9.45 & $\uparrow$ 24.05 & $\uparrow$ 8.06 & $\uparrow$ 3.96 & $\uparrow$ 16.31 & $\uparrow$ 3.46 & $\uparrow$ 0.07 & $\downarrow$ 0.66 \\
\bottomrule
\end{tabular}
\end{table*}

\subsection{Experimental Setup}
We evaluate a broad range of MLLMs across different architectures, model scales, training paradigms, and licensing settings. Our open-source baselines include Qwen2.5-VL \cite{qwen2025qwen25technicalreport} (3B, 7B, 32B, 72B), InternVL2/3 \cite{team2024internvl2,zhu2025internvl3} (8B, 38B, 78B), MiniCPM-V-2.6 \cite{yao2024minicpm}, and LLaVA-OneVision \cite{li2024llava} (7B, 72B). We further evaluate representative private models, including GPT-4o \cite{hurst2024gpt}, GPT-4.1, Gemini-2.5-Flash \cite{comanici2025gemini}, and Gemini-2.5-Pro-Thinking \cite{comanici2025gemini}. Due to space constraints, the detailed evaluation metrics for classification, detection, and VQA are provided in the Appendix D.

\subsection{Main Results}

\noindent \textbf{Classification Results.}
As shown in Table~\ref{tab:classification_results}, classification performance varies markedly between different models and scales. Among private models, GPT-4.1 achieves the best overall results. For open-source models, performance generally improves with model scale, as shown by Qwen2.5-VL increasing from 52.60\% Accuracy at 3B to 71.75\% at 72B. Notably, MARINER-7B achieves the best overall open-source performance, improving Accuracy from 64.40\% to 84.60\% and Macro F1 from 51.18\% to 83.10\% over the scale-matched Qwen2.5-VL-7B-Ins baseline, and even surpassing much larger models such as Qwen2.5-VL-72B-Ins and InternVL3-78B.

\noindent \textbf{Detection Results.}
As shown in Table~\ref{tab:detection_results}, object detection remains highly challenging, as all evaluated models obtain relatively limited performance under this setting. Among private models, Gemini-2.5-Pro achieves the best overall results, reaching 59.42\% AP and 72.68\% AR. For open-source models, larger variants generally perform better, with InternVL3-38B obtaining 27.06\% AP and the best AR of 37.05\%, while Qwen2.5-VL-72B achieves 22.49\% AP and the highest AP$_{\text{Mil}}$ of 37.29\%. In contrast, MARINER achieves 29.49\% AP, 44.32\% AP$_{50}$, 31.80\% AP$_{75}$, 36.62\% AR, and 35.93\% AP$_{\text{Civ}}$, outperforming several larger open-source models in overall detection accuracy.

\noindent \textbf{VQA Results.}
As shown in Table~\ref{tab:task_group_results}, VQA performance varies substantially across model families, indicating that realistic maritime understanding remains challenging even for strong MLLMs. Among proprietary models, Gemini-2.5-Pro and GPT-4.1 achieve the best overall results, with 73.97 and 73.77 overall accuracy, respectively. For open-source models, larger variants generally perform better, with InternVL3-38B reaching 69.29\% and Qwen2.5-VL-72B-Ins reaching 68.79\% overall. Notably, MARINER achieves the best overall open-source performance at 73.72\%, improving the scale-matched Qwen2.5-VL-7B baseline by 8.09\%.

\noindent \textbf{Human Evaluation.}
We randomly sample 500 examples from the full benchmark for a preliminary human evaluation. Human experts provide a strong reference point across classification, detection, and VQA. Overall, human performance remains competitive across all three tasks, indicating that the benchmark is well aligned with expert-level maritime understanding. However, military categories remain substantially more challenging than civilian categories for human experts, especially in fine-grained settings. This shows the intrinsic difficulty of realistic maritime in MARINER. More experimental results and analysis details are in Appendix C and D.

\subsection{Analysis.} 
Across the three tasks, several consistent patterns emerge. Proprietary models generally show stronger generalization than open-source models, especially on classification and VQA. Scaling often improves open-source performance, but the gains remain limited, particularly on the detection task. The tasks also differ clearly in difficulty: classification is relatively more manageable, detection is the most challenging, and VQA shows strong sensitivity to cognitive demand, with models performing better on perception than on spatial and reasoning.

\section{Conclusion}
We introduce MARINER, a benchmark for fine-grained maritime understanding in realistic open-water environments, built on a 3E paradigm. MARINER unifies classification, detection, and hierarchical VQA, covering fine-grained vessel categories, adverse environments, and representative incident scenarios. Extensive evaluation of mainstream MLLMs shows that, although proprietary models generally outperform open-source counterparts, current models still struggle with fine-grained recognition, robust detection, and higher-level reasoning in complex maritime scenes. These results reveal substantial limitations of current multimodal models in realistic maritime settings. We hope future work can further extend MARINER to cover a broader range of maritime tasks and support continued progress in maritime multimodal understanding.


\bibliographystyle{ACM-Reference-Format}
\bibliography{abbrev}

\appendix

\end{document}